\documentclass[sigconf]{acmart}
\AtBeginDocument{%
  }

\setcopyright{none}
\renewcommand\footnotetextcopyrightpermission[1]{}

\usepackage{multirow}

\usepackage{graphicx}
\usepackage[table,dvipsnames]{xcolor}
\definecolor{bestcolor}{RGB}{220,210,255}
\definecolor{secondcolor}{RGB}{238,232,255}
\newcommand{\tablefontsize}{\small}

\begin{document}

\title{CoRe: A Comprehensive Framework for Cross-Image Comparative Reasoning in Vision-Language Models}

\author{Lin Peng}
\orcid{0009-0008-1786-1946}
\affiliation{%
  \institution{Xi’an Jiaotong University}
  \city{Xi'an}
  \state{}
  \country{China}
}
\email{penglin@stu.xjtu.edu.cn}

\author{Cong Wan}
\orcid{0009-0007-2107-0007}
\authornote{Cong Wan is the corresponding author.}
\affiliation{%
  \institution{Xi’an Jiaotong University}
  \city{Xi'an}
  \state{}
  \country{China}
}
\email{wancong@stu.xjtu.edu.cn}

\author{Zeyu Guo}

\affiliation{%
  \institution{Xi’an Jiaotong University}
  \city{Xi'an}
  \state{}
  \country{China}
}
\email{1784279917@stu.xjtu.edu.cn}

\author{SongLin Dong}
\affiliation{%
  \institution{Xi’an Jiaotong University}
  \city{Xi'an}
  \state{}
  \country{China}
}
\email{dongsl@suat-sz.edu.cn}

\author{Yihong Gong}
\orcid{0000-0002-1793-5836}
\affiliation{%
  \institution{Xi’an Jiaotong University}
  \city{Xi'an}
  \state{}
  \country{China}
}
\email{ygong@mail.xjtu.edu.cn}


\begin{abstract}
Cross-image comparative reasoning remains challenging for vision-language models (VLMs), especially when correct prediction requires fine-grained attribute grounding and globally consistent reasoning. We present CoRe, a unified framework for this problem. CoRe includes: (i) CoRe-20K, a large-scale triplet-based training set automatically constructed from structured visual metadata through a multi-expert collaborative pipeline, covering counting, depth, distance, and spatial relations; (ii) TriSR, a structured reward framework that jointly supervises attribute grounding, judgment alignment, and triplet consistency under GRPO optimization; and (iii) CoRe-Bench, the first benchmark dedicated to fine-grained cross-image comparative reasoning. Experiments show that CoRe substantially outperforms existing VLMs on CoRe-Bench while remaining competitive on standard multimodal benchmarks, achieving a 28.2-point gain in partial accuracy over the strongest baseline.
\end{abstract}

\ccsdesc[500]{Computing methodologies~Computer vision tasks}

\keywords{Vision-language model; Cross-image comparative reasoning.}


\maketitle
\section{Introduction}

Vision-language models (VLMs)~\cite{achiam2023gpt,bai2025qwen3,chen2024internvl,liu2023visual} have achieved strong performance on standard multimodal tasks\cite{chen2025unireal,liu2025step1x,song2024processpainter,wan2024grid,peng2025cia} such as visual question answering~\cite{antol2015vqa,kuang2025natural} and image captioning~\cite{ghandi2023deep}. However, many practical scenarios~\cite{an2024etpnav,liu2024volumetric,khan2023visual,sapkota2025vision,saleh2024forest} require more than understanding each image independently: they require comparing task-relevant visual attributes across multiple related images. Such attributes may include object count, scene depth, relative distance, or spatial layout. We refer to this capability as \textit{cross-image comparative reasoning}, which arises naturally in applications such as comparing depth changes across views for navigation or count differences across scenes for monitoring.

\begin{figure*}[!t]
\centering
\includegraphics[width=1.0\textwidth]{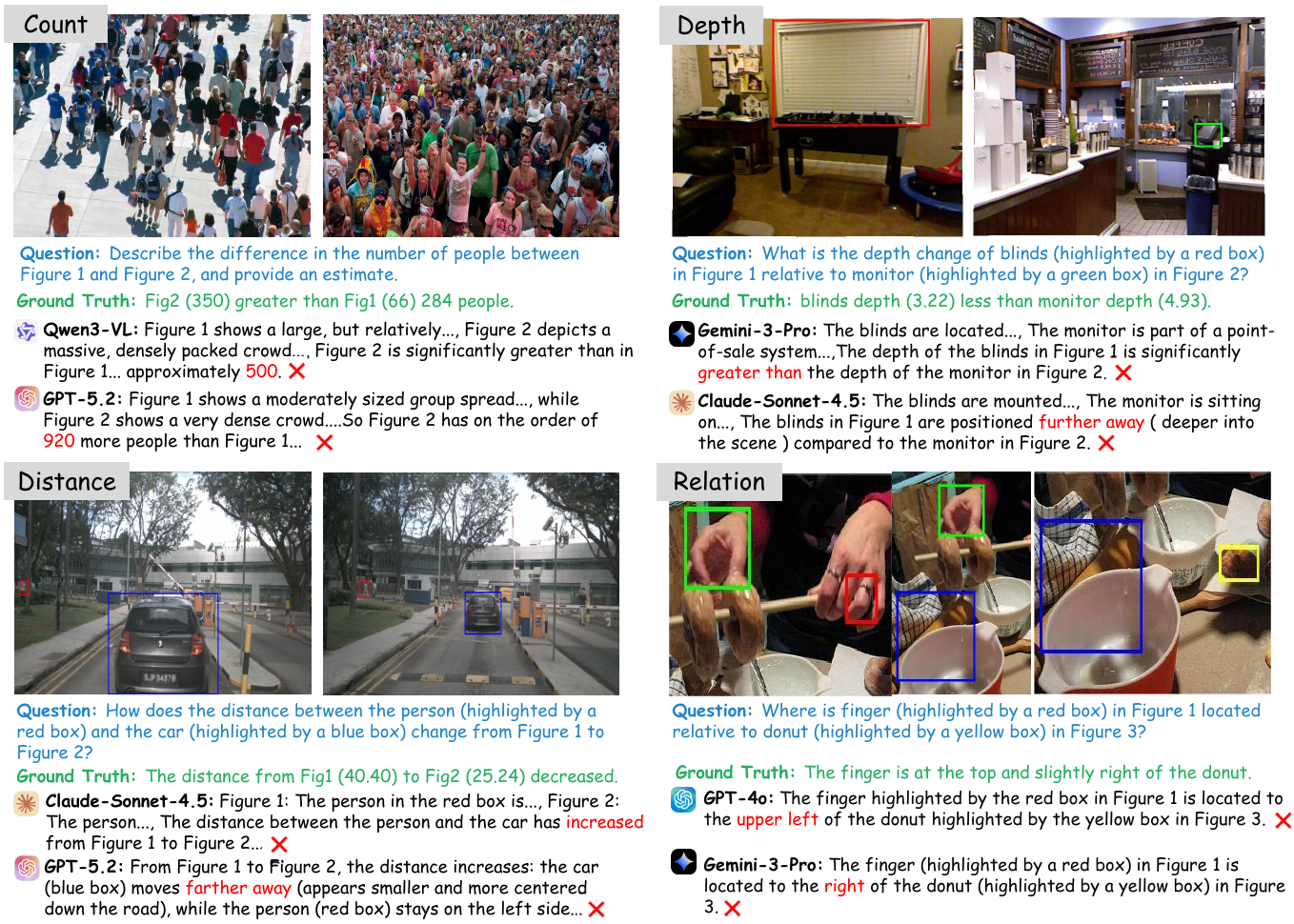}
\caption{Representative failure cases of large Vision-Language Models (VLMs) on cross-image fine-grained comparison tasks. We showcase four task categories (Count, Depth, Distance, and Relation), each demanding precise cross-image reasoning. Recent VLMs (Qwen3-VL, GPT-5.2, Claude-Sonnet-4.5, Gemini-3-Pro, and GPT-4o) consistently produce incorrect responses, exposing systematic deficiencies in existing models' cross-image visual reasoning capabilities.   }
\label{fig:vis_1}
\end{figure*}

Despite its practical relevance, cross-image comparative reasoning remains under-studied and poorly supported by current VLMs. As shown in Fig.~\ref{fig:vis_1}, even strong recent models often struggle with tasks such as cross-image counting, depth comparison, relative distance judgment, and spatial correspondence, suggesting that this is a systematic limitation rather than an isolated failure case. One reason is that existing benchmarks largely emphasize either single-image perception~\cite{tong2024cambrian,fu2024blink} or general multi-image understanding~\cite{wang2024muirbench,cheng2025evaluating}, including temporal ordering, narrative comprehension, and image retrieval. While these settings evaluate whether a model can integrate information across images at a holistic semantic level, they provide only limited coverage of fine-grained comparative reasoning, where success depends on accurately comparing specific metric attributes across images.

\begin{figure*}[!t]
\centering
\includegraphics[width=1.0\textwidth]{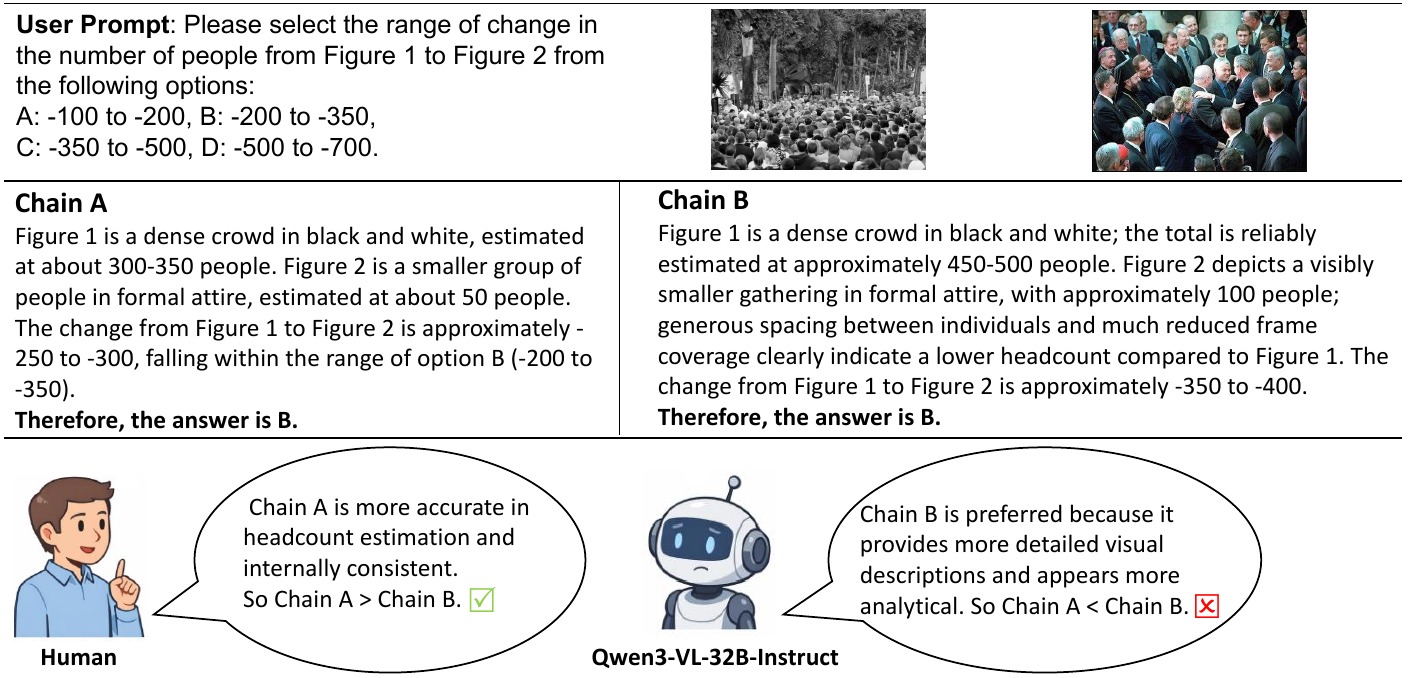}
\caption{Misalignment between LLM-based evaluation and human judgment in cross-image comparative reasoning. Although both reasoning chains reach the same final answer (B), their reasoning quality differs substantially. Human evaluators correctly prefer Chain A because it is more accurate and internally consistent, whereas the LLM judge incorrectly favors Chain B due to its richer linguistic descriptions.    }
\label{fig:cot_1}
\end{figure*}

Beyond benchmark coverage, improving this capability also poses a distinct supervision challenge. A natural strategy is to optimize models only for final-answer correctness~\cite{chen2025mico,wan2026remot,feng2025onethinker}. However, we find that outcome-only optimization is often insufficient for cross-image comparative reasoning: models may arrive at correct answers through shortcut patterns or accidental guessing while producing poorly grounded or internally inconsistent reasoning. 
In our analysis (Table~\ref{tab:pain_and_closure}), up to 17.1\% of predictions with correct final answers are accompanied by factually invalid reasoning chains, revealing a gap between answer accuracy and reasoning reliability.
A natural alternative is to use an LLM as a judge~\cite{gu2024survey,liu2025noisyrollout} to assess the reasoning trace, but this is also less suitable for our setting (Fig.~\ref{fig:cot_1}). The core issue here is not subjective preference over reasoning style, but \emph{structured reasoning correctness}: whether the model correctly identifies task-relevant attributes in each image, whether intermediate pairwise judgments are correct and consistent with the final answer, and whether all pairwise conclusions satisfy global consistency. Because these properties are explicit and verifiable, they are better handled by direct structured supervision than by holistic free-form judging.

Motivated by this observation, we present \textbf{CoRe}, a framework for studying and improving cross-image comparative reasoning in VLMs. To enable scalable and verifiable supervision, we first construct \textbf{CoRe-20K}, a large-scale triplet-based dataset in which each sample consists of three related images and their associated pairwise comparison questions. Rather than relying on manual annotation or VLM-generated pseudo-labels, CoRe-20K is built automatically from structured visual metadata through a multi-expert pipeline that extracts task-relevant metrics, filters unreliable or trivial triplets, and generates comparison questions with deterministic ground-truth labels. On top of this data, we propose \textbf{TriSR}, a structured reward framework for triplet-based reasoning that decomposes supervision into three complementary signals: \emph{Attribute Grounding}, \emph{Judgment Alignment}, and \emph{Triplet Consistency}. These rewards are optimized jointly with final-answer correctness under GRPO, encouraging models to produce not only correct answers but also grounded and globally consistent reasoning.
To evaluate this capability systematically, we further construct \textbf{CoRe-Bench}, a benchmark specifically designed for fine-grained cross-image comparative reasoning. CoRe-Bench covers four task dimensions---counting, depth, distance, and spatial relations---across diverse real-world source domains. Experiments show that CoRe substantially improves performance on this challenging setting over strong VLM baselines, while remaining competitive on broader vision-language benchmarks. Taken together, our results suggest that exploiting the verifiable intermediate structure of comparative reasoning is a promising direction for improving multi-image reasoning in VLMs.

Our contributions are three-fold: 
(1) we identify fine-grained cross-image comparative reasoning as a distinct and under-served capability of VLMs; 
(2) we propose TriSR, a structured reward framework that exploits verifiable intermediate structure in triplet-based comparison; and 
(3) we construct CoRe-20K and CoRe-Bench to enable scalable training and systematic evaluation for this setting.

\section{Related Work}

\textbf{Visual language models and evaluation benchmarks.}
Visual language models (VLMs) have advanced rapidly in recent years, evolving from early vision-language alignment frameworks such as BLIP~\cite{li2022blip} and Flamingo~\cite{alayrac2022flamingo} to instruction-following multimodal systems including LLaVA~\cite{liu2024improved} and InstructBLIP~\cite{dai2023instructblip}. More recent proprietary and open-source models, such as GPT-4o~\cite{hurst2024gpt}, Gemini~\cite{comanici2025gemini}, and Qwen3-VL~\cite{bai2025qwen3}, have further strengthened multimodal perception and reasoning. Correspondingly, a broad range of benchmarks have been proposed to assess VLM visual capabilities. Single-image benchmarks, such as MME~\cite{fu2023mme}, CV-Bench~\cite{tong2024cambrian}, MMTBench~\cite{ying2024mmt}, and BLINK~\cite{fu2024blink}, mainly evaluate perception and reasoning within individual images. More recently, multi-image benchmarks such as MuirBench~\cite{wang2024muirbench} and MMRB~\cite{cheng2025evaluating} have expanded evaluation to settings involving multiple related images, covering tasks such as retrieval, temporal ordering, and narrative understanding. However, these benchmarks primarily assess semantic integration across images at a holistic level, rather than fine-grained comparative reasoning across related images. 

\noindent\textbf{Data construction.}
Constructing reliable training data for cross-image comparative reasoning is non-trivial, as labels must capture precise metric relationships between images rather than semantic content within a single image. Existing efforts~\cite{assran2025v,liu2025spatial,dave2022tclr,sevilla2021only} typically face a trade-off between supervision quality, scalability, and label verifiability. For example, video-text datasets~\cite{bain2021frozen,grauman2022ego4d} provide only coarse clip-level descriptions, lacking the metric granularity required for explicit comparative supervision. Manual annotation can produce high-quality labels but is prohibitively labour-intensive and difficult to scale across diverse visual attributes and scene types. More recently, some approaches have used VLMs as annotators to generate pseudo-labels for comparison or reasoning data; however, this strategy risks propagating the very reasoning errors that training aims to correct, and in practice may yield high format-error rates and limited output validity~\cite{wan2026remot}. A more principled alternative is to derive comparison labels programmatically from structured metadata, which guarantees label verifiability without human annotation effort. Our CoRe-20K dataset adopts this metadata-driven strategy through a multi-expert collaborative pipeline, constructing over 20,000 high-quality triplet-based comparison samples across four metric dimensions: counting, depth, distance, and spatial relations.

\noindent\textbf{Reasoning Enhancement via Reinforcement Learning.}
The application of reinforcement learning (RL) to enhance the reasoning capabilities of large language models has attracted considerable research attention~\cite{zheng2025group,dong2025agentic,wan2026remot,chen2024measuring,shao2024visual,zhang2023multimodal,wu2025reinforcing,liu2025seg,zhang2025critique,li2025star,feng2025video,liu2025spatial}. A prominent line of work, represented by DeepSeek-R1, employs rule-based RL through the Group Relative Policy Optimization (GRPO) algorithm~\cite{guo2025deepseek}, optimizing models directly against outcome-level reward signals. While outcome-level supervision effectively drives answer accuracy, we observe that in cross-image comparative reasoning it frequently produces models that arrive at correct answers through erroneous or shortcut reasoning—a failure mode that outcome-only rewards neither diagnose nor suppress. A natural remedy is to incorporate LLM-as-Judge frameworks~\cite{gu2024survey}, which leverage powerful language models to assess intermediate reasoning quality and provide richer training signals beyond answer correctness. However, we empirically find that LLM-based evaluation exhibits a substantial alignment gap with human judgments on cross-image visual reasoning chains, rendering it unreliable as a direct reward signal for this task. This motivates our proposed TriSR framework, which decomposes holistic reasoning evaluation into three structured, verifiable criteria—Attribute Grounding, Judgment Alignment, and Triplet Consistency—transforming subjective quality assessment into a principled and reliable scoring process that yields more effective reward signals for training.

\section{Method}

\begin{figure*}[!t]
\centering
\includegraphics[width=1.0\textwidth]{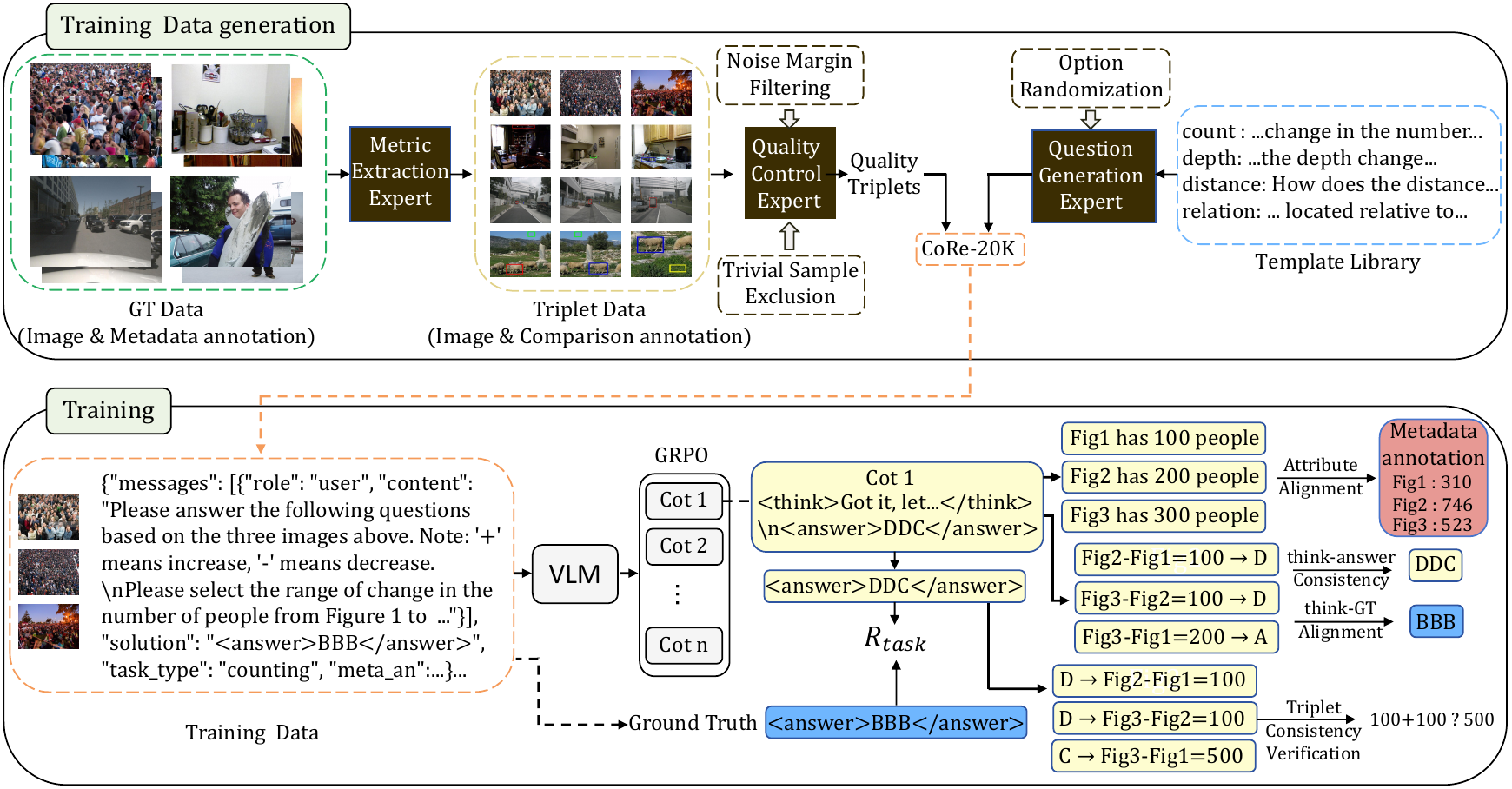}
\caption{Overview of the {CoRe} framework.(Top) Training Data Construction: A multi-expert pipeline builds CoRe-20K from structured metadata. The Metric Extraction Expert derives per-image metric values from task-specific annotations; the Quality Control Expert filters low-quality triplets via Noise Margin Filtering and Trivial Sample Exclusion; the Question Generation Expert instantiates multiple-choice questions from a Template Library with Option Randomization to eliminate answer-position bias. (Bottom) TriSR-Guided GRPO Training: The VLM generates $n$ chain-of-thought responses evaluated by a composite structured reward combining Attribute Alignment, Think--Answer Consistency, Think--GT Alignment, and Triplet Consistency Verification, which are aggregated into a group-normalized advantage to update the policy via GRPO.  }
\label{fig:cot_method}
\end{figure*}

\subsection{Problem Setup}
We study \emph{cross-image comparative reasoning} over image triplets. 
A training example is defined as
\begin{equation}\label{eq:training_example}
x = (I_1, I_2, I_3, m, \mathcal{Q}, \mathcal{A}),
\end{equation}
where \(I_1, I_2, I_3\) are three related images, \(m\) denotes the task dimension (e.g., counting, depth, distance, or spatial relation), \(\mathcal{Q}\) is the set of pairwise comparison questions, and \(\mathcal{A} = \{A_1, A_2, A_3\}\) denotes the structured metadata associated with the three images.
From the metadata, we derive a unified task-relevant attribute representation
\begin{equation}\label{eq:attribute_representation}
a_i = f_m(I_i, A_i), \qquad i \in \{1,2,3\},
\end{equation}
where \(a_i\) may be numerical (e.g., count, depth, distance) or categorical (e.g., spatial relation state), depending on the task. 
For the three image pairs
\begin{equation}\label{eq:image_pairs}
\mathcal{P} = \{(1,2), (1,3), (2,3)\},
\end{equation}
a deterministic task-specific comparator \(g_m\) produces the ground-truth pairwise labels
\begin{equation}\label{eq:ground_truth_labels}
y^*_{ij} = g_m(a_i, a_j), \qquad (i,j) \in \mathcal{P}.
\end{equation}
We denote the full target as \(y^* = \{y^*_{ij}\}_{(i,j)\in\mathcal{P}}\).

Given a triplet query \(q\), the model produces an output
\begin{equation}\label{eq:model_output}
o = (t, \hat{y}),
\end{equation}
where \(t\) is the reasoning trace and \(\hat{y}\) is the final predicted answer. 
Our goal is not only to maximize final-answer accuracy, but also to enforce that the reasoning trace is grounded in correct per-image attributes, aligned with the predicted answer, and globally consistent across the triplet.

\subsection{Metadata-Driven Triplet Construction}

\textbf{Comparison triplets.}
Unlike standard single-image instruction data, our setting requires supervision over comparisons across multiple related images. We therefore organize each sample as a \emph{comparison triplet} \(T=(I_1,I_2,I_3)\), where all three images come from the same source dataset and share a common task dimension \(m\). Each triplet is paired with three coupled comparison sub-questions over \((I_1,I_2)\), \((I_1,I_3)\), and \((I_2,I_3)\). This triplet-level formulation is crucial: it not only supports pairwise supervision, but also exposes higher-order consistency structure across the three comparisons.

\noindent\textbf{Multi-expert collaborative pipeline.}
As shown in Fig.~\ref{fig:cot_method}, we construct triplets automatically from structured metadata using a modular pipeline with three expert modules.

\noindent\textbf{Metric Extraction Expert}
\((f_m : (I_k, A_k) \rightarrow v_k)\).
Given image \(I_k\) and its task-specific annotation \(A_k\), this expert derives a unified metric representation \(v_k\). Because source datasets provide heterogeneous annotation formats, we instantiate task-specific extractors and map them into a common comparison space. Concretely, for \emph{counting}, we read instance counts from crowd annotations; for \emph{depth}, we compute the median depth of the queried region from ground-truth depth maps; for \emph{distance}, we derive ego-to-object Euclidean distance from 3D annotations; and for \emph{spatial relations}, we infer the target relation label from object-level bounding box annotations. This step decouples supervision from VLM-generated pseudo-labels and yields deterministic comparison targets.

\smallskip
\noindent\textbf{Quality Control Expert}
\((\phi_m : (v_1,v_2,v_3)\rightarrow\{0,1\})\).
Not every image triplet produces reliable or informative supervision. We therefore retain a triplet only if all three pairwise comparisons satisfy task-specific validity constraints. Let
\begin{equation}\label{eq:diff_measure}
\delta_{ij} = d_m(v_i, v_j)
\end{equation}
denote the task-specific difference measure between images \(I_i\) and \(I_j\). For numerical tasks, \(d_m\) is the absolute metric difference; for categorical tasks, it is replaced by task-specific validity rules. We accept a triplet if
\begin{equation}\label{eq:validity_cond}
\phi_m(v_1,v_2,v_3)=1
\quad \Longleftrightarrow \quad
\delta_{ij}\in\mathcal{T}_m,\ \forall (i,j)\in\mathcal{P},
\end{equation}
where \(\mathcal{T}_m\) denotes a task-adaptive admissible set.

For numerical tasks, \(\mathcal{T}_m=[\theta_{\min}^m,\theta_{\max}^m]\) serves two purposes. The lower bound \(\theta_{\min}^m\) suppresses label noise by excluding pairs whose true metric differences are too small relative to annotation uncertainty. The upper bound \(\theta_{\max}^m\) removes overly easy comparisons whose answers may be inferred from superficial cues without genuine comparative reasoning. For categorical tasks such as spatial relations, we instead enforce validity and diversity constraints to avoid degenerate triplets in which all pairwise relations collapse to the same label.

\smallskip
\noindent\textbf{Question Generation Expert.}
For each retained triplet, we instantiate a triplet-level prompt containing the three pairwise comparison sub-questions. Questions are generated from a curated template library, while the correct options are determined deterministically from the extracted metric representations. To reduce answer-position bias, we randomly shuffle the candidate options before assigning labels, ensuring that correct choices are approximately uniformly distributed across answer positions.

\noindent\textbf{CoRe-20K and CoRe-Bench.}
Using this fully automatic pipeline, we construct \textbf{CoRe-20K}, a collection of 20,000 comparison triplets with balanced coverage across four task dimensions: counting, depth, distance, and spatial relations, each contributing 5,000 triplets. Since each triplet contains three pairwise sub-questions, the full collection corresponds to 60,000 comparison QA items.

From this pool, we reserve 1,000 triplets per task dimension to form \textbf{CoRe-Bench}, yielding 4,000 held-out evaluation triplets in total. The remaining 16,000 triplets are used for training. This split preserves balanced coverage across all tasks while ensuring that evaluation is conducted on disjoint triplets.

\subsection{TriSR: Structured Reward for Triplet Reasoning}

Cross-image comparative reasoning exposes a useful property: its intermediate reasoning steps are not purely stylistic, but partially \emph{verifiable}. A correct solution should (1) identify the task-relevant attribute in each image, (2) derive pairwise judgments consistent with both the ground truth and the final answer, and (3) satisfy global consistency across the three pairwise comparisons. We therefore design \textbf{TriSR}, a structured reward framework that supervises these three aspects directly.

\textbf{Structured output parsing.}
To make intermediate supervision executable, we prompt the model to produce a structured response containing:
(1) per-image attribute estimates \(\{\hat{v}_k\}_{k=1}^3\),
(2) pairwise comparative judgments inferred in the reasoning trace \(\{\tilde{y}_{ij}\}_{(i,j)\in\mathcal{P}}\), and
(3) final predicted answers \(\{\hat{y}_{ij}\}_{(i,j)\in\mathcal{P}}\).
A lightweight rule-based parser extracts these fields from each sampled response. TriSR then assigns rewards based on the extracted structure.

\noindent\textbf{Attribute Grounding.}
The first requirement of comparative reasoning is that the model correctly grounds the task-relevant attribute in each image. We define an \emph{Attribute Grounding} reward
\begin{equation}\label{eq:reward_ag}
R_{\text{ag}}(o) = \frac{1}{3}\sum_{k=1}^{3} s_m(\hat{v}_k, v_k),
\end{equation}
where \(s_m(\cdot,\cdot)\) is a grounding score between the model-estimated attribute \(\hat{v}_k\) and the metadata-derived target \(v_k\). We use
\begin{equation}\label{eq:score_numerical}
s_m(\hat{v}_k, v_k)
=
\exp\!\left(
-\frac{|\hat{v}_k - v_k|}{\max(|v_k|,\epsilon)}
\right),
\end{equation}
where \(\epsilon>0\) avoids division by zero. 
This reward encourages the model to ground its reasoning in correct per-image attribute estimates rather than relying on ungrounded heuristics.

\noindent\textbf{Judgment Alignment.}
Even when the final answer is correct, the reasoning trace may imply the wrong pairwise conclusion, or may contradict the model's own final prediction. We therefore define a \emph{Judgment Alignment} reward that jointly measures factual correctness and internal consistency:
\begin{equation}\label{eq:reward_ja}
R_{\text{ja}}(o)
=
\frac{1}{2|\mathcal{P}|}
\sum_{(i,j)\in\mathcal{P}}
\left(
\mathbf{1}[\tilde{y}_{ij}=y^*_{ij}]
+
\mathbf{1}[\tilde{y}_{ij}=\hat{y}_{ij}]
\right).
\end{equation}
The first term checks whether the comparative judgment stated in the reasoning trace matches the ground-truth answer; the second checks whether that judgment is consistent with the model's final selected answer. Maximum reward is obtained only when the reasoning trace is both correct and self-consistent.

\noindent\textbf{Triplet Consistency.}
Pairwise comparisons over three images induce higher-order logical structure. For example, if image \(I_1\) is judged larger than \(I_2\) on a given metric and \(I_2\) is judged larger than \(I_3\), then the comparison between \(I_1\) and \(I_3\) should be globally compatible with those two intermediate judgments. We capture this property with a \emph{Triplet Consistency} reward.

We define a task-specific verifier
\begin{equation}\label{eq:verifier}
\tau_m(\hat{y}_{12}, \hat{y}_{13}, \hat{y}_{23}) \in \{-1,0,1\},
\end{equation}
where \(+1\) indicates logical consistency, \(-1\) indicates contradiction, and \(0\) denotes cases in which no non-trivial consistency check can be established.

For ordinal comparison tasks, we map each answer to a signed relation \(s_{ij}\in\{-1,0,+1\}\). If \(s_{12}\) and \(s_{23}\) jointly imply a unique relation between \(I_1\) and \(I_3\), we verify whether the predicted \(s_{13}\) matches that implication. For interval-based answers, we convert each answer to a quantitative range \(\Delta_{ij}\) and check whether the composed interval \(\Delta_{12}+\Delta_{23}\) is compatible with \(\Delta_{13}\) up to quantization tolerance. The resulting reward is
\begin{equation}\label{eq:reward_tc}
R_{\text{tc}}(o)
=
\tau_m(\hat{y}_{12}, \hat{y}_{13}, \hat{y}_{23}).
\end{equation}
Importantly, \(R_{\text{tc}}\) depends only on the model's own predictions and therefore acts as a self-consistency signal that can penalize logically incompatible outputs even when individual pairwise answers are considered in isolation.

\noindent\textbf{Composite reward.}
We combine the structured rewards with standard task accuracy. Let
\begin{equation}\label{eq:reward_task}
R_{\text{task}}(o)
=
\frac{1}{|\mathcal{P}|}
\sum_{(i,j)\in\mathcal{P}}
\mathbf{1}[\hat{y}_{ij}=y^*_{ij}]
\end{equation}
denote the average pairwise answer accuracy within a triplet. The final reward is
\begin{equation}\label{eq:reward_total}
R(o)
=
R_{\text{task}}(o)
+
\lambda_{\text{ag}} R_{\text{ag}}(o)
+
\lambda_{\text{ja}} R_{\text{ja}}(o)
+
\lambda_{\text{tc}} R_{\text{tc}}(o),
\end{equation}
where \(\lambda_{\text{ag}}, \lambda_{\text{ja}}, \lambda_{\text{tc}}\) control the relative contribution of each structured component. This formulation encourages the model to produce answers that are accurate, well-grounded, internally consistent, and globally coherent across the triplet.

\subsection{Optimization with GRPO}

We optimize the model with Group Relative Policy Optimization (GRPO)~\cite{guo2025deepseek}. For each query \(q\), we sample \(G\) responses \(\{o_i\}_{i=1}^{G}\) from the old policy \(\pi_{\theta_{\text{old}}}\), evaluate each response with the composite reward \(R(o_i)\), and compute group-normalized advantages:
\begin{equation}\label{eq:advantage}
\hat{A}_i
=
\frac{R(o_i)-\mathrm{mean}(\{R(o_j)\}_{j=1}^{G})}
{\mathrm{std}(\{R(o_j)\}_{j=1}^{G})+\epsilon}.
\end{equation}
The policy is then updated by maximizing
\begin{equation}\label{eq:objective_grpo}
J(\theta)
=
\mathbb{E}_{q,\{o_i\}}
\left[
\frac{1}{G}\sum_{i=1}^{G}
\min\!\left(
r_i \hat{A}_i,\,
\mathrm{clip}(r_i,1-\varepsilon,1+\varepsilon)\hat{A}_i
\right)
-
\beta D_{\mathrm{KL}}
\right],
\end{equation}
where
\begin{equation}\label{eq:importance_ratio}
r_i=\frac{\pi_{\theta}(o_i\mid q)}{\pi_{\theta_{\text{old}}}(o_i\mid q)}
\end{equation}
is the importance ratio, and \(\beta D_{\mathrm{KL}}\) regularizes the updated policy toward the reference distribution. Since the reward combines both outcome correctness and structured reasoning signals, GRPO encourages the model not only to obtain the correct final answer, but also to arrive there through grounded and logically consistent comparative reasoning.

\begin{table*}[t]
\centering
\caption{\textbf{Overall and partial accuracies (\%) on CoRe-Test-Bench.} We compare baseline VLMs with our trained variants across four comparative reasoning dimensions. Best results are highlighted.}
\label{tab:main}
\tablefontsize
\begin{tabular}{lcccccccccc}
\toprule
\multirow{2}{*}{\textbf{Model}} & \multicolumn{2}{c}{Count} & \multicolumn{2}{c}{Depth} & \multicolumn{2}{c}{Distance} & \multicolumn{2}{c}{Relation} & \multicolumn{2}{c}{Avg.} \\
\cmidrule(lr){2-3}\cmidrule(lr){4-5}\cmidrule(lr){6-7}\cmidrule(lr){8-9}\cmidrule(lr){10-11}
 & Ov. & Par. & Ov. & Par. & Ov. & Par. & Ov. & Par. & Ov. & Par. \\
\midrule
Qwen3-VL-4B--CoT~\cite{team2025qwen3} & 4.1 & 22.1 & 6.7 & 19.9 & 4.7 & 20.6 & 10.7 & 40.5 & 6.6 & 25.8 \\
InternVL3-2B~\cite{chen2024internvl} & 0.3 & 3.8 & 3.5 & 33.0 & 4.0 & 33.8 & 1.6 & 21.3 & 2.4 & 22.9 \\
InternVL3-8B~\cite{chen2024internvl} & 1.0 & 12.2 & 4.7 & 34.8 & 1.8 & 13.6 & 0.8 & 11.0 & 2.1 & 17.9 \\
LLaVA-OneVision-4B~\cite{li2024llava} & 1.2 & 25.3 & 5.8 & 39.8 & 7.1 & 41.9 & 0.2 & 1.5 & 3.6 & 27.1 \\
LLaVA-OneVision-8B~\cite{li2024llava} & 0.1 & 1.1 & 7.3 & 42.8 & 0.9 & 4.5 & 0.6 & 10.7 & 2.2 & 14.8 \\
SSRL-4B~\cite{liu2026spatial} & 1.4 & 21.1 & 4.3 & 35.0 & 0.8 & 4.7 & 2.2 & 17.7 & 2.2 & 19.6 \\
Remot-4B~\cite{wan2026remot} & 6.0 & 28.0 & 18.0 & 42.9 & 9.0 & 30.2 & 10.0 & 41.3 & 10.8 & 35.6 \\
\midrule
\textbf{CoRe-4B-CoT (Ours)} & \textbf{7.8} & \textbf{39.0} & \textbf{33.0} & \textbf{67.9} & \textbf{24.8} & \textbf{48.4} & \textbf{23.2} & \textbf{60.5} & \textbf{22.2} & \textbf{54.0} \\
\bottomrule
\end{tabular}
\end{table*}


\section{Experiment}
We evaluate CoRe from three complementary perspectives: (1) benchmark-level performance on CoRe-Bench, (2) diagnostic analysis of structured reasoning quality, and (3) generalization to broader VLM benchmarks.
\subsection{Experimental Setup}

\noindent\textbf{Hyperparameters.}
We adopt Qwen3-VL-4B-Thinking~\cite{bai2025qwen3} as our base model, as it provides a strong balance between reasoning capability and computational efficiency. The model is kept in Thinking mode to retain its intrinsic chain-of-thought capability. Each sample is formatted as \texttt{<think>...</think> <answer>ans</answer>}, and the task accuracy reward is computed solely against tokens within the \texttt{<answer>} block. For reinforcement learning, we adopt GRPO~\cite{guo2025deepseek} with a composite reward that combines outcome accuracy with the three proposed structured rewards, namely attribute grounding, judgment alignment, and triplet consistency. The corresponding reward weights are set to $\lambda_{ag} = 0.2$, $\lambda_{ja} = 0.2$, and $\lambda_{tc} = 0.3$. In addition, three auxiliary rewards are incorporated as standard training stabilizers: a format reward that encourages syntactically valid answer structure, a cosine CoT length reward that penalizes excessively verbose reasoning chains, and a repetition penalty that discourages degenerate token repetition. These auxiliary components are provided by the ms-swift training library and are not contributions of this work; we therefore omit their implementation details. A KL regularization coefficient of $\beta = 0.01$ constrains reward drift from the reference policy. Training uses a batch size of 4 with 4 rollouts per sample, a learning rate of $1 \times 10^{-6}$ with cosine decay schedule, and AdamW optimization, running for 1 epochs on $8\times$ A800 GPUs with mixed precision.

\noindent\textbf{Evaluation Protocols.}
We follow the default inference configuration of Qwen3-VL and evaluate all models using the VLMEvalKit toolkit~\cite{duan2024vlmevalkit}. For all benchmarks, we unify the prompting format and answer extraction rules. Each question is decoded through the model's reasoning head and parsed from the final \texttt{<answer>} token. To comprehensively capture model capability under the triplet structure, we define two complementary metrics: \textit{Overall Accuracy (Ov.)}, which marks a triplet as correct only if all three pairwise sub-questions are answered correctly, emphasizing global metric consistency; and \textit{Partial Accuracy (Par.)}, which assigns a proportional score based on the ratio of correctly answered sub-questions within a triplet, reflecting localized reasoning ability under partial understanding.

\subsection{Main Results}

\textbf{Quantitative Results.}
Table~\ref{tab:main} reports overall and partial accuracies on CoRe-Test-Bench across four comparative reasoning dimensions. Existing general-purpose VLMs perform poorly on this benchmark, with average overall accuracy remaining below 7\% for all baselines. In particular, Qwen3-VL-4B--CoT achieves 6.6\% average overall accuracy and 25.8\% partial accuracy, while InternVL3-8B and LLaVA-OneVision-8B obtain only 2.1\% and 2.2\% average overall accuracy, respectively. These results indicate that current VLMs struggle substantially with cross-image metric comparison, especially on counting and distance reasoning, where most models achieve near-zero or single-digit overall accuracy. In contrast, our \textbf{CoRe-4B-CoT} achieves \textbf{22.2\%} average overall accuracy and \textbf{54.0\%} partial accuracy, substantially outperforming all baselines. Compared with the strongest baseline, Qwen3-VL-4B--CoT, our model improves overall accuracy by \textbf{15.6} absolute points (from 6.6\% to 22.2\%) and partial accuracy by \textbf{28.2} points (from 25.8\% to 54.0\%), respectively. The improvements are consistent across all four task dimensions, with particularly large gains on \textbf{depth} (33.0\% vs.\ 6.7\%) and \textbf{distance} (24.8\% vs.\ 4.7\%), suggesting that the proposed training strategy is especially effective for cross-image comparative reasoning.

\noindent\textbf{Generalization.}
To examine whether the gains are tied to the triplet benchmark itself, we construct CoRe-OOD-Binary, a two-image out-of-distribution benchmark from external datasets that are disjoint from CoRe-Bench. As shown in Table~\ref{tab:generalization_checks}, CoRe improves over the base model on all four OOD dimensions. 


\begin{table}[t]
\centering
\caption{\textbf{Additional generalization checks.} All values are accuracies (\%). CoRe-OOD-Binary uses two-image inputs from external datasets.}
\label{tab:generalization_checks}
\tablefontsize
\setlength{\tabcolsep}{3pt}
\begin{tabular}{@{}llcc@{}}
\toprule
\textbf{Check} & \textbf{Subset} & \textbf{Qwen3-4B} & \textbf{CoRe-4B} \\
\midrule
OOD-Binary & Count (UCF-QNRF) & 26.0 & 40.0 \\
OOD-Binary & Depth (ARKitScenes) & 36.0 & 42.0 \\
OOD-Binary & Distance (Argoverse 2) & 28.0 & 53.0 \\
OOD-Binary & Relation (COCO) & 59.0 & 72.0 \\

\bottomrule
\end{tabular}
\end{table}



\begin{table}[t]
\centering
\caption{\textbf{Ablation of TriSR reward components.} Ov./Par.\ denote overall/partial accuracy (\%).}
\label{tab:reward_ablation}
\tablefontsize
\begin{tabular}{lcc}
\toprule
\textbf{Method} & \textbf{Ov.} & \textbf{Par.} \\
\midrule
Base Model (CoT, no training)                       & 6.6  & 25.8 \\
$R_{\text{task}}$ only                              & 11.5 & 43.6  \\
\midrule
TriSR w/o $R_{\text{ag}}$                           & 12.4   & 45.1   \\
TriSR w/o $R_{\text{ja}}$                           & 12.0   & 44.5   \\
TriSR w/o $R_{\text{tc}}$                           & 11.9   & 44.3   \\
\midrule
\textbf{TriSR (full)}                               & 13.1   & 45.7   \\
\bottomrule
\end{tabular}
\end{table}

\begin{table}[t]
\centering
\caption{\textbf{Core pain point quantification and solution validation.}
We decompose model outputs into four categories based on answer correctness and reasoning correctness:
\textbf{CA+CR} (correct answer + correct reasoning),
\textbf{CA+WR} (correct answer + wrong reasoning),
\textbf{WA+CR} (wrong answer + correct reasoning), and
\textbf{WA+WR} (wrong answer + wrong reasoning).
Manual inspection uses 300 triplets.}
\label{tab:pain_and_closure}
\tablefontsize
\setlength{\tabcolsep}{3pt}
\begin{tabular}{@{}lccccc@{}}
\toprule
\textbf{Model} & \textbf{Acc.} & \textbf{CA+CR} & \textbf{CA+WR} & \textbf{WA+CR} & \textbf{WA+WR} \\
\midrule
Qwen3-4B Base & 24.1 & 12.6 & 11.5 & 0.0 & 75.9 \\
Qwen3-4B + \(R_{\text{task}}\) & 42.9 & 25.8 & 17.1 & 1.1 & 56.0 \\
\midrule
\textbf{CoRe-4B} & 46.7 & 38.4 & 8.3 & 4.8 & 48.5 \\
\bottomrule
\end{tabular}
\end{table}

\begin{table*}[t]
\centering
\caption{\textbf{Evaluation on other vision-centric benchmarks.} All scores are reported in accuracy (\%). Best (\colorbox{bestcolor}{darkpurple}) and second‑best (\colorbox{secondcolor}{lightpurple}) results are highlighted.}
\label{tab:other_benchmarks}
\tablefontsize
\setlength{\tabcolsep}{5pt}
\renewcommand{\arraystretch}{1.1}
\begin{tabular}{lccccccccc}
\toprule
\textbf{Model} & \textbf{CV-Bench} & \textbf{BLINK} & \textbf{RW-QA} & \textbf{MMT} & \textbf{MMStar} & \textbf{MMVP} & \textbf{MME-RW} & \textbf{$V^*$} & \textbf{HR8K} \\
\midrule
\multicolumn{10}{l}{\textit{Proprietary Models}} \\
Claude3.7-Sonnet~\cite{duan2024vlmevalkit} & - & 56.6 & 55.4 & 60.1 & 65.1 & - & - & - & - \\
GPT-4o~\cite{hurst2024gpt} & 79.2 & 59.0 & 69.7 & - & 65.2 & 72.0 & - & 42.9 & 46.7 \\

\midrule
\multicolumn{10}{l}{\textit{Open-Source Models}} \\
Qwen2.5-VL-7B~\cite{Qwen2.5-VL} 
& 75.5 & 56.3 & 68.2 & 62.4 & 65.0 & 76.6 & 60.0 & 77.5 & 65.8 \\

LLaVA-Next-7B~\cite{liu2023improved} 
& 61.9 & 39.5 & 58.6 & 50.4 & 37.9 & 65.6 & 72.7 & 52.4 & 41.6 \\

InternVL3-2B~\cite{chen2024internvl} 
& 73.8 & 51.3 & 64.0 & 58.7 & 60.2 & 71.3 & \cellcolor{secondcolor}\textbf{84.3} & 70.7 & 57.9 \\

InternVL3-8B~\cite{chen2024internvl} 
& 83.1 & 55.3 & 70.6 & 61.9 & 66.5 & \cellcolor{bestcolor}\textbf{79.7} & \cellcolor{bestcolor}\textbf{88.7} & 68.1 & \cellcolor{bestcolor}\textbf{69.5} \\

Qwen3-VL-4B-CoT~\cite{team2025qwen3} 
& \cellcolor{secondcolor}\textbf{85.3} 
& \cellcolor{secondcolor}\textbf{59.5} 
& \cellcolor{secondcolor}\textbf{73.0} 
& \cellcolor{secondcolor}\textbf{63.4} 
& \cellcolor{secondcolor}\textbf{70.2} 
& \cellcolor{secondcolor}\textbf{79.0} 
& 71.7 
& \cellcolor{bestcolor}\textbf{79.1} 
& \cellcolor{secondcolor}\textbf{68.6} \\

\textbf{CoRe-4B-CoT (Ours)} 
& \cellcolor{bestcolor}\textbf{85.5} 
& \cellcolor{bestcolor}\textbf{60.6} 
& \cellcolor{bestcolor}\textbf{73.2} 
& \cellcolor{bestcolor}\textbf{64.5} 
& \cellcolor{bestcolor}\textbf{70.3} 
& 78.4 
& 73.8 
& \cellcolor{secondcolor}\textbf{78.6} 
& \cellcolor{bestcolor}\textbf{69.5} \\

\rowcolor[gray]{0.99}
$\Delta$ \textbf{Improvement} 
& \textcolor{ForestGreen}{+0.2} 
& \textcolor{ForestGreen}{+1.1} 
& \textcolor{ForestGreen}{+0.2} 
& \textcolor{ForestGreen}{+1.1} 
& \textcolor{ForestGreen}{+0.1} 
& {-0.6} 
& \textcolor{ForestGreen}{+2.1} 
& {-0.5} 
& \textcolor{ForestGreen}{+0.9} \\
\bottomrule
\end{tabular}
\end{table*}

\begin{figure*}[!t]
\centering
\includegraphics[width=1.0\textwidth]{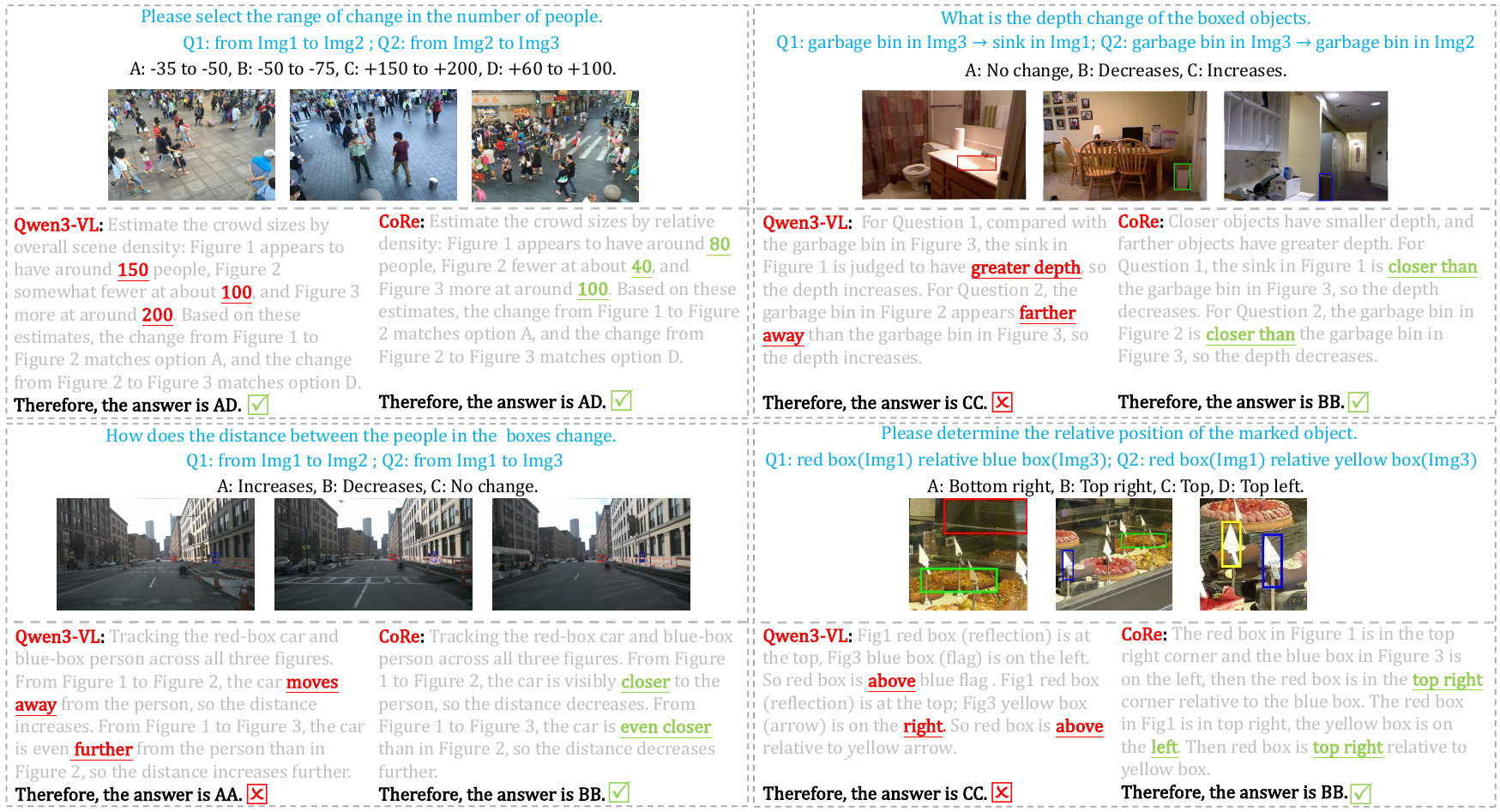}
\caption{Cross-image comparative reasoning. We compare Qwen3-VL and CoRe across four tasks: crowd counting change estimation, depth comparison, relative distance judgment, and spatial correspondence, each requiring precise fine-grained visual attribute comparison. Qwen3-VL often relies on coarse visual impressions and produces incorrect conclusions (marked with $\times$), while CoRe correctly captures fine-grained differences and yields the right answers (marked with \checkmark).}
\label{fig:cot_val}
\end{figure*}

\subsection{Ablation Study}

To analyze the contribution of each TriSR component, we conduct ablation experiments on a 10\% subset of CoRe-20K-training for efficiency; absolute accuracies are therefore lower than those in Table~\ref{tab:main}, but the relative trends remain informative.

\noindent\textbf{Effect of TriSR reward components.}
Table~\ref{tab:reward_ablation} shows that training with $R_{\text{task}}$ alone already yields substantial gains over the base model (43.6\% vs.\ 25.8\% partial accuracy), confirming that outcome-based fine-tuning provides a strong baseline. The full TriSR framework further improves partial accuracy to 45.7\%, demonstrating that structured intermediate supervision offers complementary gains beyond final-answer correctness. Ablating each component individually reveals consistent degradation: removing $R_{\text{tc}}$ causes the largest drop (45.7\%$\to$44.3\%), followed by $R_{\text{ja}}$ (44.5\%) and $R_{\text{ag}}$ (45.1\%), confirming that all three components contribute independently.

\noindent\textbf{Reasoning quality analysis.}
Table~\ref{tab:pain_and_closure} provides a finer-grained view based on manual inspection of 300 randomly sampled triplets, categorizing predictions by the joint correctness of the final answer and reasoning chain. The base model shows that nearly half of its correct predictions are accompanied by invalid reasoning, with a CA+WR rate of 11.5\% and a Wilson 95\% confidence interval of [8.4, 15.7]. 
This reveals a substantial gap between answer accuracy and reasoning reliability. Training with \(R_{\text{task}}\) alone exacerbates this issue, increasing CA+WR to 17.1\% with a Wilson 95\% confidence interval of [13.2, 21.8], indicating that outcome-only supervision reinforces shortcut reasoning. In contrast, CoRe reduces CA+WR to 8.3\% with a Wilson 95\% confidence interval of [5.7, 11.9], while increasing CA+CR from 12.6\% to 38.4\%. These results confirm that TriSR's structured rewards steer the model toward predictions that are both correct and verifiably grounded.

\subsection{Evaluation on Other VLM Benchmarks}

To assess the generalizability of CoRe beyond our proposed benchmark, we evaluate it on a diverse set of vision-centric benchmarks, including CV-Bench~\cite{tong2024cambrian}, BLINK~\cite{fu2024blink}, RealWorldQA (RW-QA)~\cite{Grok15vpreview}, MMT-Bench (MMT)~\cite{Ying2024MMTBenchAC}, MMStar~\cite{Chen2024AreWOMMStar}, MMVP~\cite{Tong2024EyesWSMMStar}, MME-RealWorld (MME-RW)~\cite{Zhang2024MMERealWorldCY}, $V^*$ Bench ($V^*$)~\cite{Wu2023VGVStar}, and HRBench (HR8K)~\cite{Wang2024DivideCAHRBench}. Results are shown in Table~\ref{tab:other_benchmarks}. CoRe demonstrates strong transferability across general VLM benchmarks and achieves the best performance among open-source models on 5 out of 9 benchmarks, including CV-Bench, BLINK, RW-QA, MMT, and MMStar. Compared with the Qwen3-VL-4B-CoT baseline, CoRe improves accuracy by 0.2\%, 1.1\%, 0.2\%, 1.1\%, and 0.1\% on these five benchmarks, respectively. CoRe also remains competitive on the remaining benchmarks, ranking second on $V^*$ and tying for the best result on HR8K. Although CoRe does not achieve the top score on MMVP and MME-RW, its performance remains strong, indicating that training for comparative reasoning does not lead to catastrophic forgetting on broader multimodal evaluation tasks. Notably, CoRe-4B matches or surpasses several larger 7B- and 8B-scale open-source models on multiple benchmarks, suggesting that targeted data construction and structured reward design can be more important than raw model scale for improving fine-grained comparative reasoning.

\subsection{Qualitative Analysis}
We visualize representative examples from CoRe-Bench in Fig.~\ref{fig:cot_val}, showing side-by-side comparisons between Qwen3-VL and CoRe on four cross-image reasoning tasks. The results show that Qwen3-VL often makes errors when comparing subtle visual differences across related images, whereas CoRe produces more reliable comparative reasoning and correct final predictions.

\textbf{Additional analyses, including training strategy comparison, triplet design validation, and limitations, are provided in the supplementary.}

\section{Conclusion}
We present CoRe, a comprehensive framework for cross-image comparative reasoning in vision-language models. Through a multi-expert collaborative pipeline, we construct CoRe-20K, a large-scale dataset derived from structured visual metadata. Combined with TriSR's structured reasoning rewards and GRPO optimization, CoRe achieves state-of-the-art performance on CoRe-Bench and multiple VLM benchmarks. Our analysis further reveals that outcome-only supervision is insufficient for this setting, as correct answers can coexist with factually invalid reasoning chains, highlighting the importance of supervising verifiable intermediate structure. 

\bibliographystyle{ACM-Reference-Format}
\bibliography{samples/sample-base}

@String{Computing = "Computing" }

@String{Computer = "{IEEE} Computer" }

@String{Springer = "Springer-Verlag" }

@article{guo2025deepseek,
  title={Deepseek-r1: Incentivizing reasoning capability in llms via reinforcement learning},
  author={Guo, Daya and Yang, Dejian and Zhang, Haowei and Song, Junxiao and Wang, Peiyi and Zhu, Qihao and Xu, Runxin and Zhang, Ruoyu and Ma, Shirong and Bi, Xiao and others},
  journal={arXiv preprint arXiv:2501.12948},
  year={2025}
}

@inproceedings{antol2015vqa,
  title={Vqa: Visual question answering},
  author={Antol, Stanislaw and Agrawal, Aishwarya and Lu, Jiasen and Mitchell, Margaret and Batra, Dhruv and Zitnick, C Lawrence and Parikh, Devi},
  booktitle={Proceedings of the IEEE international conference on computer vision},
  pages={2425--2433},
  year={2015}
}

@article{kuang2025natural,
  title={Natural language understanding and inference with mllm in visual question answering: A survey},
  author={Kuang, Jiayi and Shen, Ying and Xie, Jingyou and Luo, Haohao and Xu, Zhe and Li, Ronghao and Li, Yinghui and Cheng, Xianfeng and Lin, Xika and Han, Yu},
  journal={ACM Computing Surveys},
  volume={57},
  number={8},
  pages={1--36},
  year={2025},
  publisher={ACM New York, NY}
}

@article{ghandi2023deep,
  title={Deep learning approaches on image captioning: A review},
  author={Ghandi, Taraneh and Pourreza, Hamidreza and Mahyar, Hamidreza},
  journal={ACM Computing Surveys},
  volume={56},
  number={3},
  pages={1--39},
  year={2023},
  publisher={ACM New York, NY}
}

@article{sapkota2025vision,
  title={Vision-Language-Action (VLA) Models: Concepts, Progress, Applications and Challenges},
  author={Sapkota, Ranjan and Cao, Yang and Roumeliotis, Konstantinos I and Karkee, Manoj},
  journal={arXiv preprint arXiv:2505.04769},
  year={2025}
}

@article{an2024etpnav,
  title={Etpnav: Evolving topological planning for vision-language navigation in continuous environments},
  author={An, Dong and Wang, Hanqing and Wang, Wenguan and Wang, Zun and Huang, Yan and He, Keji and Wang, Liang},
  journal={IEEE Transactions on Pattern Analysis and Machine Intelligence},
  year={2024},
  publisher={IEEE}
}

@inproceedings{liu2024volumetric,
  title={Volumetric environment representation for vision-language navigation},
  author={Liu, Rui and Wang, Wenguan and Yang, Yi},
  booktitle={Proceedings of the IEEE/CVF conference on computer vision and pattern recognition},
  pages={16317--16328},
  year={2024}
}

@article{khan2023visual,
  title={Visual crowd analysis: Open research problems},
  author={Khan, Muhammad Asif and Menouar, Hamid and Hamila, Ridha},
  journal={AI Magazine},
  volume={44},
  number={3},
  pages={296--311},
  year={2023},
  publisher={Wiley Online Library}
}

@article{saleh2024forest,
  title={Forest fire surveillance systems: A review of deep learning methods},
  author={Saleh, Azlan and Zulkifley, Mohd Asyraf and Harun, Hazimah Haspi and Gaudreault, Francis and Davison, Ian and Spraggon, Martin},
  journal={Heliyon},
  volume={10},
  number={1},
  year={2024},
  publisher={Elsevier}
}

@article{tong2024cambrian,
  title={Cambrian-1: A fully open, vision-centric exploration of multimodal llms},
  author={Tong, Shengbang and Brown, Ellis and Wu, Penghao and Woo, Sanghyun and Middepogu, Manoj and Akula, Sai C and Yang, Jihan and Yang, Shusheng and Iyer, Adithya and Pan, Xichen and others},
  journal={Advances in Neural Information Processing Systems},
  volume={37},
  pages={87310--87356},
  year={2024}
}

@article{fu2023mme,
  title={Mme: A comprehensive evaluation benchmark for multimodal large language models},
  author={Fu, Chaoyou and Chen, Peixian and Shen, Yunhang and Qin, Yulei and Zhang, Mengdan and Lin, Xu and Yang, Jinrui and Zheng, Xiawu and Li, Ke and Sun, Xing and others},
  journal={arXiv preprint arXiv:2306.13394},
  year={2023}
}

@article{achiam2023gpt,
  title={Gpt-4 technical report},
  author={Achiam, Josh and Adler, Steven and Agarwal, Sandhini and Ahmad, Lama and Akkaya, Ilge and Aleman, Florencia Leoni and Almeida, Diogo and Altenschmidt, Janko and Altman, Sam and Anadkat, Shyamal and others},
  journal={arXiv preprint arXiv:2303.08774},
  year={2023}
}

@article{bai2025qwen3,
  title={Qwen3-vl technical report},
  author={Bai, Shuai and Cai, Yuxuan and Chen, Ruizhe and Chen, Keqin and Chen, Xionghui and Cheng, Zesen and Deng, Lianghao and Ding, Wei and Gao, Chang and Ge, Chunjiang and others},
  journal={arXiv preprint arXiv:2511.21631},
  year={2025}
}

@inproceedings{chen2024internvl,
  title={Internvl: Scaling up vision foundation models and aligning for generic visual-linguistic tasks},
  author={Chen, Zhe and Wu, Jiannan and Wang, Wenhai and Su, Weijie and Chen, Guo and Xing, Sen and Zhong, Muyan and Zhang, Qinglong and Zhu, Xizhou and Lu, Lewei and others},
  booktitle={Proceedings of the IEEE/CVF conference on computer vision and pattern recognition},
  pages={24185--24198},
  year={2024}
}

@article{hurst2024gpt,
  title={Gpt-4o system card},
  author={Hurst, Aaron and Lerer, Adam and Goucher, Adam P and Perelman, Adam and Ramesh, Aditya and Clark, Aidan and Ostrow, AJ and Welihinda, Akila and Hayes, Alan and Radford, Alec and others},
  journal={arXiv preprint arXiv:2410.21276},
  year={2024}
}

@article{liu2023visual,
  title={Visual instruction tuning},
  author={Liu, Haotian and Li, Chunyuan and Wu, Qingyang and Lee, Yong Jae},
  journal={Advances in neural information processing systems},
  volume={36},
  pages={34892--34916},
  year={2023}
}

@inproceedings{li2022blip,
  title={Blip: Bootstrapping language-image pre-training for unified vision-language understanding and generation},
  author={Li, Junnan and Li, Dongxu and Xiong, Caiming and Hoi, Steven},
  booktitle={International conference on machine learning},
  pages={12888--12900},
  year={2022},
  organization={PMLR}
}

@article{alayrac2022flamingo,
  title={Flamingo: a visual language model for few-shot learning},
  author={Alayrac, Jean-Baptiste and Donahue, Jeff and Luc, Pauline and Miech, Antoine and Barr, Iain and Hasson, Yana and Lenc, Karel and Mensch, Arthur and Millican, Katherine and Reynolds, Malcolm and others},
  journal={Advances in neural information processing systems},
  volume={35},
  pages={23716--23736},
  year={2022}
}

@inproceedings{liu2024improved,
  title={Improved baselines with visual instruction tuning},
  author={Liu, Haotian and Li, Chunyuan and Li, Yuheng and Lee, Yong Jae},
  booktitle={Proceedings of the IEEE/CVF conference on computer vision and pattern recognition},
  pages={26296--26306},
  year={2024}
}

@article{dai2023instructblip,
  title={Instructblip: Towards general-purpose vision-language models with instruction tuning},
  author={Dai, Wenliang and Li, Junnan and Li, Dongxu and Tiong, Anthony and Zhao, Junqi and Wang, Weisheng and Li, Boyang and Fung, Pascale N and Hoi, Steven},
  journal={Advances in neural information processing systems},
  volume={36},
  pages={49250--49267},
  year={2023}
}

@article{comanici2025gemini,
  title={Gemini 2.5: Pushing the frontier with advanced reasoning, multimodality, long context, and next generation agentic capabilities},
  author={Comanici, Gheorghe and Bieber, Eric and Schaekermann, Mike and Pasupat, Ice and Sachdeva, Noveen and Dhillon, Inderjit and Blistein, Marcel and Ram, Ori and Zhang, Dan and Rosen, Evan and others},
  journal={arXiv preprint arXiv:2507.06261},
  year={2025}
}

@article{ying2024mmt,
  title={Mmt-bench: A comprehensive multimodal benchmark for evaluating large vision-language models towards multitask agi},
  author={Ying, Kaining and Meng, Fanqing and Wang, Jin and Li, Zhiqian and Lin, Han and Yang, Yue and Zhang, Hao and Zhang, Wenbo and Lin, Yuqi and Liu, Shuo and others},
  journal={arXiv preprint arXiv:2404.16006},
  year={2024}
}

@inproceedings{fu2024blink,
  title={Blink: Multimodal large language models can see but not perceive},
  author={Fu, Xingyu and Hu, Yushi and Li, Bangzheng and Feng, Yu and Wang, Haoyu and Lin, Xudong and Roth, Dan and Smith, Noah A and Ma, Wei-Chiu and Krishna, Ranjay},
  booktitle={European Conference on Computer Vision},
  pages={148--166},
  year={2024},
  organization={Springer}
}

@article{zheng2025group,
  title={Group sequence policy optimization},
  author={Zheng, Chujie and Liu, Shixuan and Li, Mingze and Chen, Xiong-Hui and Yu, Bowen and Gao, Chang and Dang, Kai and Liu, Yuqiong and Men, Rui and Yang, An and others},
  journal={arXiv preprint arXiv:2507.18071},
  year={2025}
}

@article{dong2025agentic,
  title={Agentic reinforced policy optimization},
  author={Dong, Guanting and Mao, Hangyu and Ma, Kai and Bao, Licheng and Chen, Yifei and Wang, Zhongyuan and Chen, Zhongxia and Du, Jiazhen and Wang, Huiyang and Zhang, Fuzheng and others},
  journal={arXiv preprint arXiv:2507.19849},
  year={2025}
}

@article{feng2025onethinker,
  title={Onethinker: All-in-one reasoning model for image and video},
  author={Feng, Kaituo and Zhang, Manyuan and Li, Hongyu and Fan, Kaixuan and Chen, Shuang and Jiang, Yilei and Zheng, Dian and Sun, Peiwen and Zhang, Yiyuan and Sun, Haoze and others},
  journal={arXiv preprint arXiv:2512.03043},
  year={2025}
}

@article{wan2026remot,
  title={ReMoT: Reinforcement Learning with Motion Contrast Triplets},
  author={Wan, Cong and Guo, Zeyu and Li, Jiangyang and Dong, SongLin and Bai, Yifan and Peng, Lin and Ma, Zhiheng and Gong, Yihong},
  journal={arXiv preprint arXiv:2603.00461},
  year={2026}
}

@article{gu2024survey,
  title={A survey on llm-as-a-judge},
  author={Gu, Jiawei and Jiang, Xuhui and Shi, Zhichao and Tan, Hexiang and Zhai, Xuehao and Xu, Chengjin and Li, Wei and Shen, Yinghan and Ma, Shengjie and Liu, Honghao and others},
  journal={The Innovation},
  year={2024},
  publisher={Elsevier}
}

@inproceedings{duan2024vlmevalkit,
  title={Vlmevalkit: An open-source toolkit for evaluating large multi-modality models},
  author={Duan, Haodong and Yang, Junming and Qiao, Yuxuan and Fang, Xinyu and Chen, Lin and Liu, Yuan and Dong, Xiaoyi and Zang, Yuhang and Zhang, Pan and Wang, Jiaqi and others},
  booktitle={Proceedings of the 32nd ACM international conference on multimedia},
  pages={11198--11201},
  year={2024}
}

@article{li2024llava,
  title={Llava-onevision: Easy visual task transfer},
  author={Li, Bo and Zhang, Yuanhan and Guo, Dong and Zhang, Renrui and Li, Feng and Zhang, Hao and Zhang, Kaichen and Zhang, Peiyuan and Li, Yanwei and Liu, Ziwei and others},
  journal={arXiv preprint arXiv:2408.03326},
  year={2024}
}

@article{team2025qwen3,
  title={Qwen3-vl: Sharper vision, deeper thought, broader action},
  author={Team, Qwen},
  journal={Qwen Blog. Accessed},
  pages={10--04},
  year={2025}
}

@blog{Grok15vpreview,
  title={Grok-1.5 Vision Preview},
  author={XAI},
  year={2024},
  url={https://x.ai/news/grok-1.5v}
}

@article{Ying2024MMTBenchAC,
  title={MMT-Bench: A Comprehensive Multimodal Benchmark for Evaluating Large Vision-Language Models Towards Multitask AGI},
  author={Kaining Ying and Fanqing Meng and Jin Wang and Zhiqiang Li and Han Lin and Yue Yang and Hao Zhang and Wenbo Zhang and Yuqi Lin and Shuo Liu and Jiayi Lei and Quanfeng Lu and Runjian Chen and Peng Xu and Renrui Zhang and Haozhe Zhang and Peng Gao and Yali Wang and Yuning Qiao and Ping Luo and Kaipeng Zhang and Wenqi Shao},
  journal={ArXiv},
  year={2024},
  volume={abs/2404.16006},
  url={https://api.semanticscholar.org/CorpusID:269362969}
}

@article{Wang2024DivideCAHRBench,
  title={Divide, Conquer and Combine: A Training-Free Framework for High-Resolution Image Perception in Multimodal Large Language Models},
  author={Wenbin Wang and Liang Ding and Minyan Zeng and Xiabin Zhou and Li Shen and Yong Luo and Dacheng Tao},
  journal={ArXiv},
  year={2024},
  volume={abs/2408.15556},
  url={https://api.semanticscholar.org/CorpusID:271974926}
}

@article{Wu2023VGVStar,
  title={V*: Guided Visual Search as a Core Mechanism in Multimodal LLMs},
  author={Penghao Wu and Saining Xie},
  journal={2024 IEEE/CVF Conference on Computer Vision and Pattern Recognition (CVPR)},
  year={2023},
  pages={13084-13094},
  url={https://api.semanticscholar.org/CorpusID:266436019}
}

@article{Zhang2024MMERealWorldCY,
  title={MME-RealWorld: Could Your Multimodal LLM Challenge High-Resolution Real-World Scenarios that are Difficult for Humans?},
  author={Yi-Fan Zhang and Huanyu Zhang and Haochen Tian and Chaoyou Fu and Shuangqing Zhang and Jun Wu and Feng Li and Kun Wang and Qingsong Wen and Zhang Zhang and Liang Wang and Rong Jin and Tien-Ping Tan},
  journal={ArXiv},
  year={2024},
  volume={abs/2408.13257},
  url={https://api.semanticscholar.org/CorpusID:271947320}
}

@article{Tong2024EyesWSMMStar,
  title={Eyes Wide Shut? Exploring the Visual Shortcomings of Multimodal LLMs},
  author={Shengbang Tong and Zhuang Liu and Yuexiang Zhai and Yi Ma and Yann LeCun and Saining Xie},
  journal={2024 IEEE/CVF Conference on Computer Vision and Pattern Recognition (CVPR)},
  year={2024},
  pages={9568-9578},
  url={https://api.semanticscholar.org/CorpusID:266976992}
}

@article{Chen2024AreWOMMStar,
  title={Are We on the Right Way for Evaluating Large Vision-Language Models?},
  author={Lin Chen and Jinsong Li and Xiao-wen Dong and Pan Zhang and Yuhang Zang and Zehui Chen and Haodong Duan and Jiaqi Wang and Yu Qiao and Dahua Lin and Feng Zhao},
  journal={ArXiv},
  year={2024},
  volume={abs/2403.20330},
  url={https://api.semanticscholar.org/CorpusID:268793433}
}

@article{Qwen2.5-VL,
  title={Qwen2.5-VL Technical Report},
  author={Bai, Shuai and Chen, Keqin and Liu, Xuejing and Wang, Jialin and Ge, Wenbin and Song, Sibo and Dang, Kai and Wang, Peng and Wang, Shijie and Tang, Jun and Zhong, Humen and Zhu, Yuanzhi and Yang, Mingkun and Li, Zhaohai and Wan, Jianqiang and Wang, Pengfei and Ding, Wei and Fu, Zheren and Xu, Yiheng and Ye, Jiabo and Zhang, Xi and Xie, Tianbao and Cheng, Zesen and Zhang, Hang and Yang, Zhibo and Xu, Haiyang and Lin, Junyang},
  journal={arXiv preprint arXiv:2502.13923},
  year={2025}
}

@misc{liu2023improved,
      title={Improved Baselines with Visual Instruction Tuning}, 
      author={Haotian Liu and Chunyuan Li and Yuheng Li and Yong Jae Lee},
      year={2023},
      eprint={2310.03744},
      archivePrefix={arXiv},
      primaryClass={cs.CV}
}

@article{wang2024muirbench,
  title={Muirbench: A comprehensive benchmark for robust multi-image understanding},
  author={Wang, Fei and Fu, Xingyu and Huang, James Y and Li, Zekun and Liu, Qin and Liu, Xiaogeng and Ma, Mingyu Derek and Xu, Nan and Zhou, Wenxuan and Zhang, Kai and others},
  journal={arXiv preprint arXiv:2406.09411},
  year={2024}
}

@inproceedings{bain2021frozen,
  title={Frozen in time: A joint video and image encoder for end-to-end retrieval},
  author={Bain, Max and Nagrani, Arsha and Varol, G{\"u}l and Zisserman, Andrew},
  booktitle={Proceedings of the IEEE/CVF international conference on computer vision},
  pages={1728--1738},
  year={2021}
}

@inproceedings{grauman2022ego4d,
  title={Ego4d: Around the world in 3,000 hours of egocentric video},
  author={Grauman, Kristen and Westbury, Andrew and Byrne, Eugene and Chavis, Zachary and Furnari, Antonino and Girdhar, Rohit and Hamburger, Jackson and Jiang, Hao and Liu, Miao and Liu, Xingyu and others},
  booktitle={Proceedings of the IEEE/CVF conference on computer vision and pattern recognition},
  pages={18995--19012},
  year={2022}
}

@article{chen2025mico,
  title={Mico: Multi-image contrast for reinforcement visual reasoning},
  author={Chen, Xi and Zhu, Mingkang and Liu, Shaoteng and Wu, Xiaoyang and Xu, Xiaogang and Liu, Yu and Bai, Xiang and Zhao, Hengshuang},
  journal={arXiv preprint arXiv:2506.22434},
  year={2025}
}

@inproceedings{chen2024measuring,
  title={Measuring and improving chain-of-thought reasoning in vision-language models},
  author={Chen, Yangyi and Sikka, Karan and Cogswell, Michael and Ji, Heng and Divakaran, Ajay},
  booktitle={Proceedings of the 2024 Conference of the North American Chapter of the Association for Computational Linguistics: Human Language Technologies (Volume 1: Long Papers)},
  pages={192--210},
  year={2024}
}

@article{shao2024visual,
  title={Visual cot: Advancing multi-modal language models with a comprehensive dataset and benchmark for chain-of-thought reasoning},
  author={Shao, Hao and Qian, Shengju and Xiao, Han and Song, Guanglu and Zong, Zhuofan and Wang, Letian and Liu, Yu and Li, Hongsheng},
  journal={Advances in Neural Information Processing Systems},
  volume={37},
  pages={8612--8642},
  year={2024}
}

@article{zhang2023multimodal,
  title={Multimodal chain-of-thought reasoning in language models},
  author={Zhang, Zhuosheng and Zhang, Aston and Li, Mu and Zhao, Hai and Karypis, George and Smola, Alex},
  journal={arXiv preprint arXiv:2302.00923},
  year={2023}
}

@inproceedings{chen2025unireal,
  title={Unireal: Universal image generation and editing via learning real-world dynamics},
  author={Chen, Xi and Zhang, Zhifei and Zhang, He and Zhou, Yuqian and Kim, Soo Ye and Liu, Qing and Li, Yijun and Zhang, Jianming and Zhao, Nanxuan and Wang, Yilin and others},
  booktitle={Proceedings of the Computer Vision and Pattern Recognition Conference},
  pages={12501--12511},
  year={2025}
}

@article{liu2025step1x,
  title={Step1x-edit: A practical framework for general image editing},
  author={Liu, Shiyu and Han, Yucheng and Xing, Peng and Yin, Fukun and Wang, Rui and Cheng, Wei and Liao, Jiaqi and Wang, Yingming and Fu, Honghao and Han, Chunrui and others},
  journal={arXiv preprint arXiv:2504.17761},
  year={2025}
}

@article{song2024processpainter,
  title={Processpainter: Learn painting process from sequence data},
  author={Song, Yiren and Huang, Shijie and Yao, Chen and Ye, Xiaojun and Ci, Hai and Liu, Jiaming and Zhang, Yuxuan and Shou, Mike Zheng},
  journal={arXiv preprint arXiv:2406.06062},
  year={2024}
}

@article{wan2024grid,
  title={Grid: Visual layout generation},
  author={Wan, Cong and Luo, Xiangyang and Cai, Zijian and Song, Yiren and Zhao, Yunlong and Bai, Yifan and He, Yuhang and Gong, Yihong},
  journal={arXiv e-prints},
  pages={arXiv--2412},
  year={2024}
}

@inproceedings{peng2025cia,
  title={CIA: Class-and Instance-aware Adaptation for Vision-Language Models},
  author={Peng, Lin and Wan, Cong and Wang, Shaokun and Song, Xiang and He, Yuhang and Gong, Yihong},
  booktitle={Proceedings of the 33rd ACM International Conference on Multimedia},
  pages={2870--2879},
  year={2025}
}

@article{wu2025reinforcing,
  title={Reinforcing spatial reasoning in vision-language models with interwoven thinking and visual drawing},
  author={Wu, Junfei and Guan, Jian and Feng, Kaituo and Liu, Qiang and Wu, Shu and Wang, Liang and Wu, Wei and Tan, Tieniu},
  journal={arXiv preprint arXiv:2506.09965},
  year={2025}
}

@article{liu2025seg,
  title={Seg-zero: Reasoning-chain guided segmentation via cognitive reinforcement},
  author={Liu, Yuqi and Peng, Bohao and Zhong, Zhisheng and Yue, Zihao and Lu, Fanbin and Yu, Bei and Jia, Jiaya},
  journal={arXiv preprint arXiv:2503.06520},
  year={2025}
}

@article{zhang2025critique,
  title={Critique-grpo: Advancing llm reasoning with natural language and numerical feedback},
  author={Zhang, Xiaoying and Zhang, Yipeng and Sun, Hao and Feng, Kaituo and Lu, Chaochao and Yang, Chao and Meng, Helen},
  journal={arXiv preprint arXiv:2506.03106},
  year={2025}
}

@article{li2025star,
  title={Star-r1: Spatial transformation reasoning by reinforcing multimodal llms},
  author={Li, Zongzhao and Ma, Zongyang and Li, Mingze and Li, Songyou and Rong, Yu and Xu, Tingyang and Zhang, Ziqi and Zhao, Deli and Huang, Wenbing},
  journal={arXiv preprint arXiv:2505.15804},
  year={2025}
}

@article{feng2025video,
  title={Video-r1: Reinforcing video reasoning in mllms},
  author={Feng, Kaituo and Gong, Kaixiong and Li, Bohao and Guo, Zonghao and Wang, Yibing and Peng, Tianshuo and Wu, Junfei and Zhang, Xiaoying and Wang, Benyou and Yue, Xiangyu},
  journal={arXiv preprint arXiv:2503.21776},
  year={2025}
}

@article{cheng2025evaluating,
  title={Evaluating mllms with multimodal multi-image reasoning benchmark},
  author={Cheng, Ziming and Xu, Binrui and Gong, Lisheng and Song, Zuhe and Zhou, Tianshuo and Zhong, Shiqi and Ren, Siyu and Chen, Mingxiang and Meng, Xiangchao and Zhang, Yuxin and others},
  journal={arXiv preprint arXiv:2506.04280},
  year={2025}
}

@article{liu2025noisyrollout,
  title={Noisyrollout: Reinforcing visual reasoning with data augmentation},
  author={Liu, Xiangyan and Ni, Jinjie and Wu, Zijian and Du, Chao and Dou, Longxu and Wang, Haonan and Pang, Tianyu and Shieh, Michael Qizhe},
  journal={arXiv preprint arXiv:2504.13055},
  year={2025}
}

@article{liu2025spatial,
  title={Spatial-ssrl: Enhancing spatial understanding via self-supervised reinforcement learning},
  author={Liu, Yuhong and Zhang, Beichen and Zang, Yuhang and Cao, Yuhang and Xing, Long and Dong, Xiaoyi and Duan, Haodong and Lin, Dahua and Wang, Jiaqi},
  journal={arXiv preprint arXiv:2510.27606},
  year={2025}
}

@article{assran2025v,
  title={V-jepa 2: Self-supervised video models enable understanding, prediction and planning},
  author={Assran, Mido and Bardes, Adrien and Fan, David and Garrido, Quentin and Howes, Russell and Muckley, Matthew and Rizvi, Ammar and Roberts, Claire and Sinha, Koustuv and Zholus, Artem and others},
  journal={arXiv preprint arXiv:2506.09985},
  year={2025}
}

@article{dave2022tclr,
  title={Tclr: Temporal contrastive learning for video representation},
  author={Dave, Ishan and Gupta, Rohit and Rizve, Mamshad Nayeem and Shah, Mubarak},
  journal={Computer Vision and Image Understanding},
  volume={219},
  pages={103406},
  year={2022},
  publisher={Elsevier}
}

@inproceedings{sevilla2021only,
  title={Only time can tell: Discovering temporal data for temporal modeling},
  author={Sevilla-Lara, Laura and Zha, Shengxin and Yan, Zhicheng and Goswami, Vedanuj and Feiszli, Matt and Torresani, Lorenzo},
  booktitle={Proceedings of the IEEE/CVF winter conference on applications of computer vision},
  pages={535--544},
  year={2021}
}

@inproceedings{liu2026spatial,
  title={Spatial-ssrl: Enhancing spatial understanding via self-supervised reinforcement learning},
  author={Liu, Yuhong and Zhang, Beichen and Zang, Yuhang and Cao, Yuhang and Xing, Long and Dong, Xiaoyi and Duan, Haodong and Lin, Dahua and Wang, Jiaqi},
  booktitle={Proceedings of the IEEE/CVF Conference on Computer Vision and Pattern Recognition},
  pages={9570--9581},
  year={2026}
}


\end{document}